\documentclass[11pt]{article}
\usepackage{coling2020}
\usepackage{times}
\usepackage{url}
\usepackage{latexsym}

\usepackage{CJKutf8}
\usepackage{float}
\usepackage{microtype}
\usepackage{multirow}
\usepackage{booktabs}
\usepackage{amsmath}
\usepackage{amsfonts,amssymb}
\usepackage{makecell}
\usepackage{subfigure}
\usepackage{graphicx}
\usepackage{enumitem}
\usepackage{amsfonts}
\usepackage{array}
\usepackage{CJKutf8}
\usepackage{balance}
\usepackage[ruled,linesnumbered]{algorithm2e}

\colingfinalcopy 

\title{Personalized Multimodal Feedback Generation in Education}

\author{Haochen Liu\textsuperscript{1}, Zitao Liu\textsuperscript{2}\thanks{\hspace{0.1cm} The corresponding author: Zitao Liu.}, Zhongqin Wu\textsuperscript{2} and Jiliang Tang\textsuperscript{1} \\
\textsuperscript{\rm 1}  Michigan State University, East Lansing, MI, USA \\
\textsuperscript{\rm 2} TAL Education Group, Beijing, China \\
  {\tt liuhaoc1@msu.edu, liuzitao@100tal.com,}\\
  {\tt wuzhongqin@100tal.com, tangjili@msu.edu}\\}

\date{}

\begin{document}
\maketitle
\begin{abstract}
The automatic feedback of school assignments is an important application of AI in education. In this work, we focus on the task of personalized multimodal feedback generation, which aims to generate personalized feedback for teachers to evaluate students' assignments involving multimodal inputs such as images, audios, and texts. This task involves the representation and fusion of multimodal information and natural language generation, which presents the challenges from three aspects: (1) how to encode and integrate multimodal inputs; (2) how to generate feedback specific to each modality; and (3) how to fulfill personalized feedback generation. In this paper, we propose a novel \textbf{Personalized Multimodal Feedback Generation Network} (PMFGN) armed with a modality gate mechanism and a personalized bias mechanism to address these challenges. Extensive experiments on real-world K-12 education data show that our model significantly outperforms  baselines by generating more accurate and diverse feedback. In addition, detailed ablation experiments are conducted to deepen our understanding of the proposed framework.
\end{abstract}

\section{Introduction}

\begin{figure}[t]
\centering
\includegraphics[width=\linewidth]{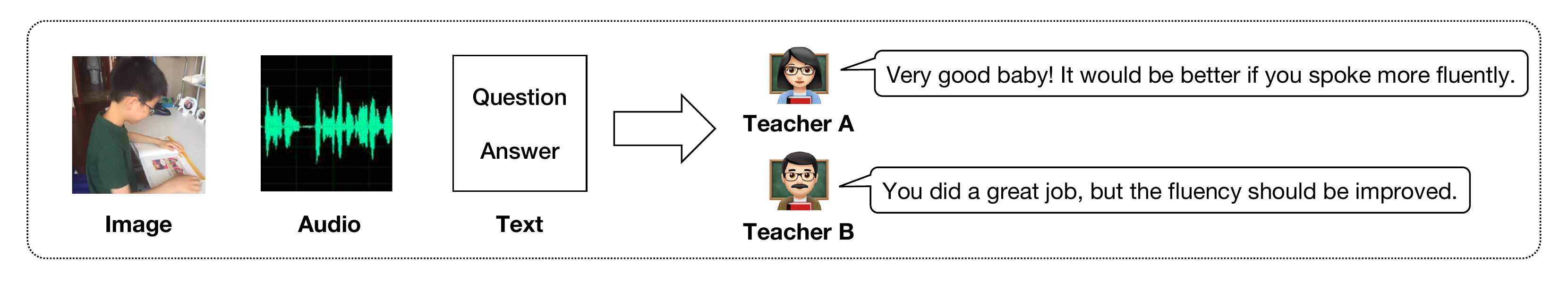}
\caption{An example of the multimodal feedback generation task. Given the same multimodal inputs,  teachers may provide different feedback.}
\label{fig:example}
\end{figure}

In recent years, oral presentations have become a popular form of assignments in online K-12 education \cite{liu2020recent}. An oral presentation assignment requires students to answer a question or explain a concept verbally. It is able to test students' oral expression skills, language organization abilities, and understanding of the topic itself simultaneously \cite{liu2019automatic}. Oral assignments are usually submitted in video format, which involves multimodal information. For example, given a math question ``Please describe the procedure of finding the greatest common divisor of two integers'', a student is asked to record a video in which he or she presents the answer to teachers. The video contains information on the modality of image (pictures of the video), audio (voice of the student), and text (transcribed speech). To evaluate such oral presentation assignments, teachers on the online education platform provide short textual feedback. An example is shown in Figure \ref{fig:example}. Feedback may involve different aspects of modalities, such as the clarity of video, the fluency of voice, and the relevance between the answers and the questions. Meanwhile, different teachers tend to write feedback on their own language styles. In order to lighten the workloads of  teachers and improve the efficiency of online teaching, in this paper, we aim to automatically generate personalized feedback from multimodal oral presentation assignments based on the language styles of various teachers. The task, known as personalized multimodal feedback generation, is an important but rarely touched application of AI in Education \cite{liu2020dolphin}.

One traditional solution of this task is to construct a pipeline-based system. Firstly, raw information from different modalities is passed into a series of pre-trained models for image recognition, speech fluency evaluation, and text relevance assessment. Then, according to some manually designed strategies, a piece of artificial feedback is retrieved from a repository. The main limitations of the pipeline methods are: (1) the terminal supervisory signals (the feedback texts in our case) cannot be propagated to upstream modules. Thus we have to use extra labeled data to train them, also, domain knowledge may be required; (2) different modules depend on each other, therefore the errors from upstream modules may directly lead to the errors in downstream modules \cite{DBLP:journals/corr/ZhaoE16}; and (3) it is hard to achieve personalization if we only rely on a limited feedback repository. 

To address the above shortcomings of traditional solution and build a more realistic end-to-end educational feedback generation system, we face unique challenges. First, since the input is composed of information from multiple modalities, it is challenging to represent and fuse multimodal information for feedback generation \cite{DBLP:journals/pami/BaltrusaitisAM19}. Moreover, to help students better understand the feedback, the generated content should be specific to each modality. In addition, teachers have different styles, and it is desired to imitate the language styles of various teachers and generate personalized feedback.

To address the above challenges, in this work, we propose a novel deep learning architecture named Personalized Multimodal Feedback Generation Network (PMFGN) which can be trained in an end-to-end manner. Although there exist some end-to-end multimodal-based text generation approaches \cite{DBLP:journals/corr/KirosSZ14,DBLP:journals/corr/MaoXYWY14a,DBLP:conf/iccv/HoriHLZHHMS17}, most of them cannot handle inputs with the form of image, audio, and text simultaneously. To the best of our knowledge, our model with a novel network structure is the first one designed for the task of multimodal feedback generation. In our proposed framework, we introduce a modality gate with a hierarchical attention mechanism that enables the model to generate different parts of the feedback based on the evaluation of different modalities. Meanwhile, we design a personalized bias mechanism to encourage the model to generate personalized feedback. Experiments have been conducted on a real-world K-12 oral presentation dataset. The results have verified the superiority of our proposed model according to various evaluation metrics. Compared with several baselines, our method can provide more precise, reasonable, and diverse feedback. We summarize our major contributions as follows:

\begin{itemize}
    \item We propose a modality gate with a hierarchical attention mechanism to enable multimodal data integration and feedback generation to specific modalities.
    \item We introduce a novel personalized bias mechanism to realize personalized feedback generation.
    \item We build a novel PMFGN model for the multimodal feedback generation task. It is shown to achieve state-of-the-art performance on this task by experiments.
\end{itemize}

The rest of the paper is organized as follows. First, we introduce the task definition of personalized multimodal feedback generation in Section \ref{sec:task}. Afterward, we present our PMFGN framework in Section \ref{sec:model}. Next, Section \ref{sec:exp} carries out our experimental setup and results with discussions. Then, we review related works in Section \ref{sec:rela}. Finally, Section \ref{sec:con} concludes the work with possible future research directions.
\section{Task Definition}
\label{sec:task}

Typically the feedback given by teachers is mainly concerned about the clarity of the video, the fluency of the voice, and the relevance between the answer and the given topic. Thus, given a question $q$ and a video submission, we extract raw information from three modalities: image, audio, and text. Since images don't change much from beginning to end when a student giving an oral presentation, we just take a screenshot of the video, i.e. an image $i$, as the signals of image modality. We separate the sound from the video as the audio signals $a$. Besides, we transcribe the audio by automated speech recognition (ASR) tools into texts $t$. Both the question content $q$ and the ASR transcription $t$ compose the signals of textual modality.

We formulate the task of personalized multimodal feedback generation as follows. Given a corpus $X=\{(i,a,t,q,p,y)\}_{n=1}^{N}$, where each instance contains an image $i$, a piece of audio $a$, a speech text $t$, a question text $q$, a teacher identifier $p$ and the corresponding feedback text $y$ written by that teacher, we seek to train a model which can generate personalized feedback for the teacher $p$ given a set of the above multimodal inputs $(i,a,t,q)$.
\section{The Proposed Model}
\label{sec:model}

\subsection{Overview}

\begin{figure*}[!t]
\includegraphics[width=\linewidth]{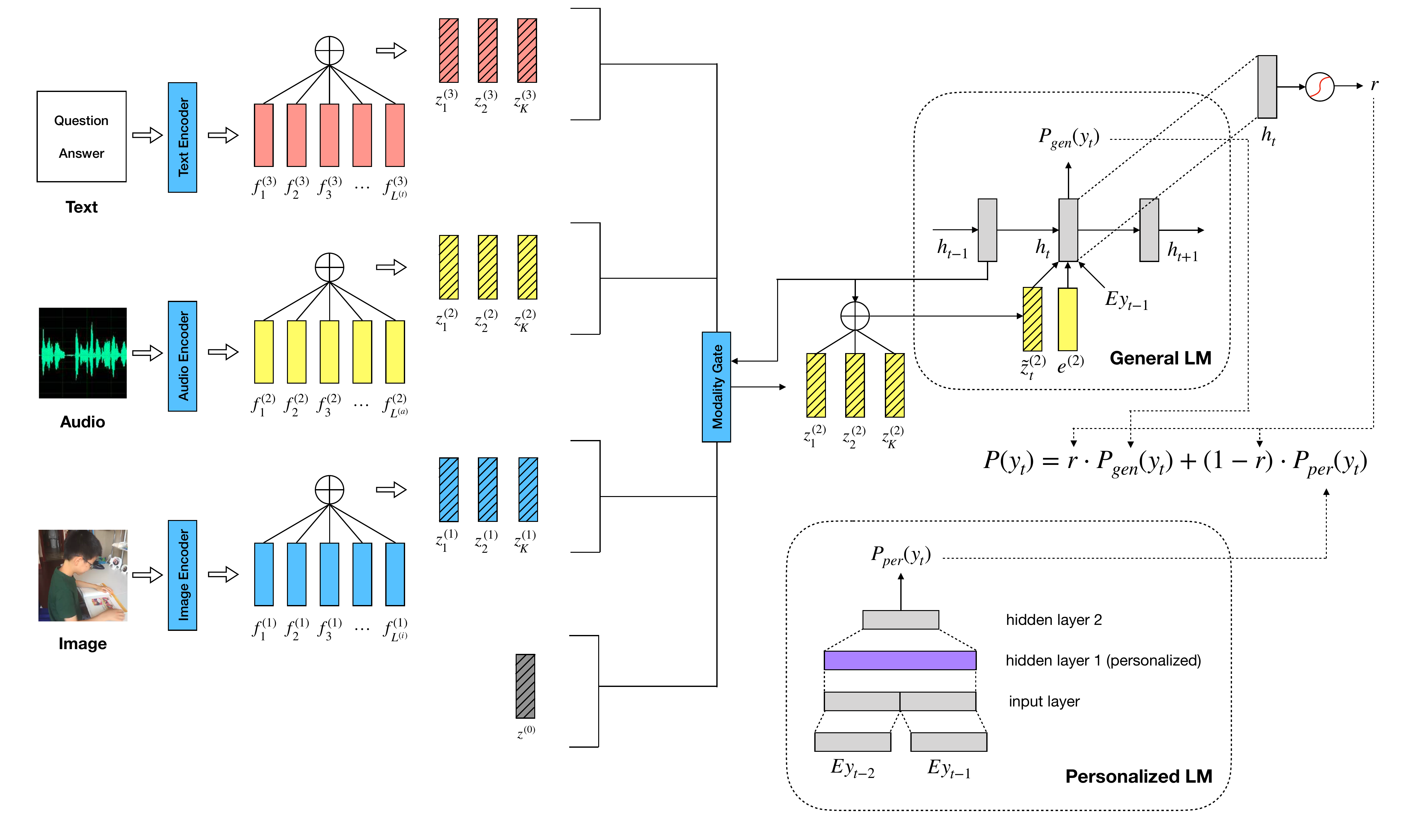}
\caption{The overview of the framework. The illustration shows an example where the audio information is selected to pass through the modality gate at the current decoding step.}
\label{fig:model}
\end{figure*}

The overall framework of the model is presented in Figure \ref{fig:model}. The model consists of four components to address the aforementioned challenges: (1) modality encoders, which encode the raw image, audio, and text inputs into a sequence of feature representations; (2) modality gate, which selects information to generate the feedback specific to each modality; (3) general language model, which is an RNN decoder that models the distribution of the feedback conditioned on the encoded modality information; and (4) personalized language model, which provides a bias on the distribution estimated by the general language model in order to imitate the wording and tone of  individual teachers. The framework is an end-to-end architecture that takes the image, audio, and texts as input and generates the feedback as output.

\subsection{Modality Encoders}

\noindent \textbf{Image Encoder.} The image encoder transforms an image into a sequence of vector representations with fixed-length $L^{(i)}$, each of which stands for a specific area of that image. We use GoogLeNet to extract these image features \cite{DBLP:conf/cvpr/SzegedyLJSRAEVR15}:
\begin{equation}
    \mathbf{F^{(1)}}=(\mathbf{f_1^{(1)}}, \mathbf{f_2^{(1)}}, \dots, \mathbf{f_{L^{(i)}}^{(1)}}) = {\rm \textbf{CNN}}(i)\nonumber
\end{equation}

\noindent \textbf{Audio Encoder.} We first extract a sequence of acoustic features, i.e., Mel-Frequency Cepstral Coefficients (MFCCs), $\mathbf{a_1}, \mathbf{a_2}, \dots, \mathbf{a_{L^{(a)}}}$ from the original audio signals, and then encode them through an unidirectional Recurrent Neural Network, where the hidden state of each timestamp is treated as the audio features of this timestamp:

\begin{equation}
    \mathbf{F^{(2)}}=(\mathbf{f_1^{(2)}}, \mathbf{f_2^{(2)}}, \dots, \mathbf{f_{L^{(a)}}^{(2)}}) = {\rm \textbf{RNN}}(\mathbf{a_1}, \mathbf{a_2}, \dots, \mathbf{a_{L^{(a)}}})\nonumber
\end{equation}
\noindent Vanilla RNN cell, GRU, or LSTM can be used as the recurrent cells.

\noindent \textbf{Text Encoder.} To facilitate the whole framework to learn to estimate the relevance between the speech text $t$ and the question text $q$, we use a MatchRNN structure \cite{DBLP:conf/naacl/WangJ16} to encode them simultaneously. Given a speech text $t=(t_1, t_2, \dots, t_{L^{(t)}})$ with length $L^{(t)}$ and a question text $q=(q_1, q_2, \dots, q_{L^{(q)}})$ with length $L^{(q)}$, we pass them through two RNNs and obtain the corresponding hidden state sequences $\mathbf{H^t}=(\mathbf{h_1^t}, \mathbf{h_2^t}, \dots, \mathbf{h_{L^{(t)}}^t})$, $\mathbf{H^q}=(\mathbf{h_1^q}, \mathbf{h_2^q}, \dots, \mathbf{h_{L^{(q)}}^q)}$.
Then for each word $t_k$ in the speech text, we compute an attention-weighted combinations of the hidden states of the question text:
\begin{align}
    \label{eq:ck}
    \mathbf{c_k}=\sum_{j=1}^{L^{(q)}} \alpha_{kj}\mathbf{h_j^q}, \qquad k=1, \dots,L^{(t)}
\end{align}

Afterward, the attentive vector $\mathbf{c_k}$ and the hidden state $\mathbf{h_k^t}$ of the speech text are concatenated as $\mathbf{m_k}=[\mathbf{c_k}:\mathbf{h_k^t}]$, which is then fed into a matching RNN with LSTM cells: $\mathbf{h_k^m}={\rm LSTM}(\mathbf{h_{k-1}^m}, \mathbf{m_k})$. The obtained hidden state $\mathbf{h_k^m}$ implies the matching degree between each word $t_k$ in the speech text and the question text $q$. More important matching results are selectively remembered. In Equation \ref{eq:ck}, the attention weight is determined by $\alpha_{kj}={\rm softmax}(e_{kj})$, and $e_{kj}$ is defined as $e_{kj}=\mathbf{w} \cdot \tanh{(\mathbf{W^q} \mathbf{h_j^q} + \mathbf{W^t} \mathbf{h_k^t} +  \mathbf{W^m} \mathbf{h _{k-1}^m})}$, where $\mathbf{w}$ is a parameter vector and $\mathbf{W^q}$, $\mathbf{W^t}$, $\mathbf{W^m}$ are parameter matrices.

Unlike the original MatchRNN, which only takes the last hidden state $\mathbf{h_{L^{(t)}}^m}$ for prediction, our framework treats all the hidden states $(\mathbf{h_1^m}, \mathbf{h_2^m}, \dots, \mathbf{h_{L^{(t)}}^m})$ as the encoded features for the downstream feedback generation task. Finally, we get the matching features of the speech text and the question text:
\begin{equation}
    \mathbf{F^{(3)}}=(\mathbf{f_1^{(3)}}, \mathbf{f_2^{(3)}}, \dots, \mathbf{f_{L^{(t)}}^{(3)}}) =(\mathbf{h_1^m}, \mathbf{h_2^m}, \dots, \mathbf{h_{L^{(t)}}^m})\nonumber
\end{equation}

All the feature vectors of three modalities are set to be $d$-dimensional. As a unified network, the three modality encoders are trained together with the rest parts in an end-to-end manner.

\subsection{Modality Gate}

After encoding the information of multiple modalities, how to effectively integrate such information for downstream tasks is one of the most important challenges \cite{DBLP:journals/pami/BaltrusaitisAM19}. In the feedback generation task, teachers may include specific comments on the information of various modalities and general comments on the entire submission in one piece of feedback. Thus, to meet the special needs of this task, a reasonable way to integrate multimodal information is, at each step of decoding, we choose the information of one noteworthy modality based on the context to generate a word in the feedback. We introduce a modality gate mechanism which selectively allows the information of one modality to pass through at each step for the decoder to generate the feedback. When there is no modality that should receive special attention, a pre-defined feature vector indicating ``general comment'' will pass through the gate to generate such feedback.

What's more, for the information of one modality, there may exist various aspects of evaluation. Let's take the modality of audio as an example. Defective audio may suffer from several different flaws like the volume is too low, the voice is not fluent or the voice is unclear. Thus, before deciding which modality to focus on, we perform structured self-attention \cite{DBLP:conf/iclr/LinFSYXZB17} with $K$ hops on the modality features, to represent the quality of this modality in different aspects (say, the quality of the audio in volume, fluency, clarity) as a weighted sum of the modality features. Ideally, the result of each hop stands for an aspect. For modality $m$, the representation of the $k$-th aspect is calculated as
\begin{equation}
    \mathbf{z_k^{(m)}}=\sum_{j=1}^{L^{(m)}} \alpha_{jk}\mathbf{f_j^{(m)}}, k=1,2,\dots,K\nonumber
\end{equation}

\noindent where $\alpha_{kj}={\rm softmax}(e_{kj})$ and $e_{kj}=\mathbf{q_k^{(m)}} \tanh{(\mathbf{W^{(m)}}\mathbf{f_j^{(m)}})}$. Here, $\mathbf{q_k^{(m)}} \in \mathbf{R}^d$ is a learnable query vector and $\mathbf{W^{(m)}} \in \mathbf{R}^{d \times d}$ is a learnable parameter matrix. Besides the aspect vectors $\{\mathbf{z_k^{(1)}}\}_{k=1}^K$, $\{\mathbf{z_k^{(2)}}\}_{k=1}^K$ and $\{\mathbf{z_k^{(3)}}\}_{k=1}^K$ of the three modalities, we introduce a single learnable vector $\mathbf{z^{(0)}} \in \mathbb{R}^d$ to indicate the ``general comment''.

In the modality gate, we first map the four groups of $\mathbf{z}$ vectors to key vectors for four-class classification to decide whether to generate a general comment or a comment targeted on image, audio or texts. We use $\mathbf{z^{(0)}}$ as its key identically $\mathbf{k^{(0)}}=\mathbf{z^{(0)}}$, and map the aspect vectors $\{\mathbf{z_k^{(m)}}\}_{k=1}^K$ to their keys $\mathbf{k^{(m)}} \in \mathbb{R}^d$ by concatenating them together and performing a linear transformation $\mathbf{k^{(m)}}=\mathbf{W_k^{(m)}}[\mathbf{z_1^{(m)}}:\dots:\mathbf{z_K^{(m)}}]$, $m=1,2,3$, where $\mathbf{W_k^{(m)}} \in \mathbb{R}^{d \times Kd}$ is a parameter matrix. Given the keys $\mathbf{k}=[\mathbf{k^{(0)}}, \mathbf{k^{(1)}}, \mathbf{k^{(2)}}, \mathbf{k^{(3)}}]$, we use the hidden state of the decoder at the previous step $\mathbf{h_{t-1}}$ as the query to determine which modality to pass. Their corresponding scores $\mathbf{s} \in \mathbb{R}^{4}$ are calculated as follows:
\begin{equation}
    \mathbf{s}=\mathrm{softmax}(\mathbf{h_{t-1}}\cdot\tanh{(\mathbf{Wk})})\nonumber
\end{equation}
\noindent where $\mathbf{W} \in \mathbb{R}^{d \times d}$ is a learnable parameter matrix.



During training, we use signals in the feedback to indicate which modality each sentence is directed at. For example, when only voice-related phrases like ``couldn't hear'', ``not fluent'' appear in a sentence, we will assign a label of ``audio'' for each word in that sentence. Only the $\mathbf{z}$-vectors of the corresponding modality pass through the modality gate to generate the words in the sentence. The modality labels are provided as supervisory signals to train the classifier in the modality gate, by optimizing the following cross-entropy objective function:

\begin{equation}
    J'=- \frac{1}{N}\sum_{i=1}^{N}(\frac{1}{L_i}\sum_{l=1}^{L_i}\sum_{m=0}^{3} y_{il}^m \log s_{il}^m)\nonumber
\end{equation}
\noindent where $N$ is the number of the training instances, $L_i$ is the length of the feedback, $y_{il}^m$ is 1 when the $l$-th word belongs to modality $m$, and 0 otherwise. $s_{il}^m$ indicates the likelihood predicted by the modality gate that it belongs to modality $m$. This objective function will be trained jointly with the loss function in Equation \ref{equ:obj2} in the form of multi-task learning \cite{DBLP:conf/icml/Caruana93}.

\subsection{General Language Model}
A conditioned GRU language model is used as the decoder to generate the feedback conditioned on the $\mathbf{z}$-vectors that pass through the modality gate. In contrast to the personalized language model below, we refer to it as the general language model.

For the modality $m$, at the step $t$, we align the previous hidden state $\mathbf{h_{t-1}}$ with the aspect vectors $\{\mathbf{z_k^{(m)}}\}_{k=1}^K$ to compute a weighted sum of them, so that we can enable the model to focus on different aspects when generating different words:
\begin{equation}
    \mathbf{\tilde{z}_t^{(m)}} = \sum_{k=1}^{K} \alpha_{tk} \mathbf{z_k^{(m)}} \qquad
    \alpha_{tk}=\mathrm{softmax}(e_{tk}) \qquad
    e_{tk}=\mathbf{h_{t-1}}\cdot \tanh{(\mathbf{W_h}\mathbf{z_k^{(m)}})}\nonumber
\end{equation}

\noindent where $\mathbf{W_h} \in \mathbb{R}^{d_h \times d}$ is a parameter matrix. When $m$ is 0, we keep $\mathbf{\tilde{z}_t^{(0)}}$ as a constant vector $\mathbf{\tilde{z}_t^{(0)}} \equiv \mathbf{z^{(0)}}$.

For each modality $m$, we introduce a learnable embedding vector $\mathbf{e^{(m)}} \in \mathbb{R}^d$ to indicate which modality is targeted on for generating the feedback. Then the word at the previous step $y_{t-1}$, the modality embedding $\mathbf{e^{(m)}}$, together with $\mathbf{\tilde{z}_t^{(m)}}$, are taken as the inputs of the RNN to update the hidden state:
\begin{equation}
    \mathbf{h_t}=\mathrm{GRU}(\mathbf{h_{t-1}}, \mathbf{E}y_{t-1}, \mathbf{e^{(m)}},\mathbf{\tilde{z}_t^{(m)}})\nonumber
\end{equation}
\noindent where $\mathbf{E}$ is the embedding matrix of the words.

The probability of the next word $y_t$ predicted by the general language model is computed by an output layer after the hidden states:
\begin{equation}
    P_{gen}(y_t|y_1,\dots,y_{t-1},\mathbf{e^{(m)}},\mathbf{\tilde{z}_t^{(m)})})=g_{gen}(\mathbf{h_t})\nonumber
\end{equation}
\noindent where $g_{gen}$ represents the function of the output layer in the general language model.

\subsection{Personalized Language Model}
The general language model estimates the distribution of the feedback conditioned on the encoded modality information. It is trained on a great amount of feedback data from all the teachers so it tends to generate generic feedback. To address the challenge of personalized feedback generation, we introduce a personalized language model that models another distribution of the feedback conditioned on the teacher. The latter performs as a bias towards the former to help the model generate feedback in a specific style of an individual teacher.

We apply a bigram DNN-based language model as the personalized language model. Taking the embeddings of the previous two words $\mathbf{x}=[\mathbf{E}y_{t-2}:\mathbf{E}y_{t-1}]$ as input, the network passes them through two hidden layers and then predicts the probability of the current word $y_t$:
\begin{align}
    \mathbf{h}=\mathbf{H_2}\cdot\tanh{(\mathbf{H_1^{p}x} +\mathbf{d_1^p})}+\mathbf{d_2} \qquad
    P_{per}(y_t)=g_{per}(\mathbf{h})\nonumber
\end{align}

\noindent where $g_{per}$ represents the function of the output layer in the personalized language model. Unlike the original model, we use distinct parameters $\mathbf{H_1^{p}}$ and $\mathbf{d_1^p}$ in the first hidden layer for each teacher to enable the model to learn the different language styles of different teachers. 

\subsection{Loss Function}

The final distribution of the word $y_t$ is computed as a linear combination of the probability predicted by the general language model and the personalized language model:
\begin{equation}
    P(y_t)=r \cdot P_{gen}(y_t) + (1-r) \cdot P_{per}(y_t) \qquad
    r=\sigma(\mathbf{w_r}\cdot\mathbf{h_t})\nonumber
\end{equation}
\noindent where the weight $r$ is decided by the current hidden state $\mathbf{h_t}$ of the general language model, $\sigma$ is the sigmoid function and $\mathbf{w_r} \in \mathbb{R}^{d_h}$ is a parameter vector. In this way, the model can learn to automatically judge whether we should write a word based on the modality information or the teacher's personal preferences at each step.

The model is trained to minimize the negative log-likelihood of the ground truth feedback:
\begin{equation}
    J=-\sum_{i=1}^{N}\sum_{t=1}^{L}\log P(y_t)
    \label{equ:obj2}
\end{equation}
Hence, the final joint loss function is
\begin{equation}
    L(\Theta)=J+\alpha J'\nonumber
\end{equation}
\noindent where $\alpha$ is a hyperparameter that adjusts the weighted balance of two parts.
\section{Experiment}
\label{sec:exp}
In this section, we conduct extensive experiments to evaluate our proposed framework based on a real-world dataset collected from an online education platform. Through the experiments, we try to answer two questions: (1) Does our model achieve better performances than representative baselines? and (2) How does each component in our proposed framework contribute to the performance?


\subsection{Dataset}
The Dolphin dataset contains $20,442$ videos of oral presentation assignments collected from a real-world online education platform. The assignments are about answering a math question at the level of kindergarten or primary school. Each video is accompanied by a piece of feedback manually written by a teacher. A total of $111$ different teachers wrote the feedback. The average length of the feedback is $8.78$ Chinese characters. The dataset is randomly divided into $15,550$ records for training, $1,945$ for validation and $2,947$ for test. We guarantee that in each set, the data of every teacher are selected in proportion to the total number of that teacher's data.


\subsection{Implementation Details}
In this subsection, we describe the details of the implementation of our proposed model. Speech texts, question texts, and feedback texts are first segmented by Jieba Chinese segmentation system\footnote{http://pypi.python.org/pypi/jieba}. We build two distinct dictionaries for the texts from the input (i.e. speech texts and question texts) and the output (i.e. feedback texts). We initialize the word embeddings from standard normal distribution $N(0,1)$. All the other parameters are initialized from a uniform distribution $U(-1/\sqrt{k}, 1/\sqrt{k})$, where $k$ is the size of the last dimension of the parameter tensor.

Hyper-parameters are determined according to the model performance on the validation set as follows. We adopt the pre-trained GoogLeNet (Inception v1) model \cite{DBLP:conf/cvpr/SzegedyLJSRAEVR15} as the image encoder, where the layers behind inception (5b) are removed and replaced with a linear layer that transforms the features to the dimension of $256$. The audio data are first downsampled to 16 kHz, and then we extract the MFCCs of it from 50 ms time windows with a 50 ms shift. The audio encoder is implemented as a GRU RNN with a hidden size of $256$. For both the text encoder and the general language model, we set the size of word embeddings as $256$ and the size of the hidden states as $512$. GRU cells are used for the general language model. We perform structured self-attention with $K=3$ hops on all three modalities. The value of $\alpha$ is chosen as $0.5$. An Adam optimizer \cite{kingma2014adam} with an initial learning rate of $0.001$ is applied to train the model. The batch size is set to be $10$. When the perplexity of the model on the validation data doesn't drop for 3 consecutive epochs, the training is terminated.

\begin{table*}[t]
\centering
  \caption{Performance comparison in terms of automatic metrics. Dist-1,2 represents the average scores of Distinct-1 and Distinct-2.}
  \vspace{5pt}
  \begin{tabular}{l|c|c|c|c|c|c}
  \hline
    \textbf{Models} & \textbf{Perplexity} & \textbf{BLEU-1} & \textbf{BLEU-2} & \textbf{BLEU-3} & \textbf{ROUGE} & \textbf{Dist-1,2 (\%)}\\
    \hline
    \textbf{GRU-LM} & 15.557 & 0.2294 & 0.1392 & 0.0998 & 0.1873 & 2.19\\
    \textbf{Show-Attend-and-Tell} & 14.823 & 0.3825 & 0.2582 & 0.1880 & 0.3370 & 0.11\\
    \textbf{Attribute2Seq-Audio} & 19.066 & 0.4050 & 0.3274 & 0.2798 & 0.3563 & 0.61\\
    \textbf{Attribute2Seq-Text} & 14.901 & 0.4239 & 0.3621 & 0.3287 & 0.3652 & 0.63\\
    \textbf{Multimodal Attention} & 15.229 & 0.4505 & 0.3702 & 0.3251 & 0.3832 & 0.52\\
    \textbf{Repository} & / & 0.1849 & 0.0738 & 0.0350 & 0.2032 & 0.70\\
    \textbf{PMFGN} & \textbf{10.158} & \textbf{0.5159} & \textbf{0.4453} & \textbf{0.4063} & \textbf{0.5096} & \textbf{2.80}\\
  \hline
  \end{tabular}
  \label{tab:expresults}
\end{table*}

\subsection{Baselines}
We compare our model with the following methods.
\begin{itemize}
    \item \textbf{GRU-LM}: A GRU language model trained on the feedback in the training set (i.e., the alone general language model in our framework). In the test phase, we randomly sample the first word and then generate an entire piece of feedback by greedy search.
    \item \textbf{Show-Attend-and-Tell} \cite{DBLP:conf/icml/XuBKCCSZB15}: Originally it is an image captioning model with an attention mechanism. We use it to generate feedback based on the image features.
    \item \textbf{Attribute2Seq-Audio} \cite{DBLP:conf/eacl/ZhouLWDHX17}: The model generates product reviews based on a sequence of attribute vectors with an attention mechanism. We substitute the audio features for the attributes to generate the feedback for assignments.
    \item \textbf{Attribute2Seq-Text}: It is similar to Attribute2Seq-Audio but it takes text features as input.
    \item \textbf{Multimodal Attention} \cite{DBLP:conf/iccv/HoriHLZHHMS17}: The model adopts a hierarchical attention mechanism to fuse multimodal information to generate video descriptions. We use it to integrate image, audio and text features for feedback generation.
    \item \textbf{Repository}: We also include a repository-based baseline. It is the traditional pipeline method that selects a piece of feedback from several repositories according to a simple strategy. First, we use three well-trained binary classification models to judge whether there is no sound in the audio, whether the voice of the student is fluent, and whether the speech text and the question text are relevant. And then for each of the four cases: ``no sound'', ``not relevant'', ``relevant but not fluent'', ``relevant and fluent'', we select a piece of feedback from the corresponding feedback repository. The four repositories above contain $34$, $29$, $96$, $92$ pieces of feedback written by different teachers respectively. The sound existence model and the fluence model are realized by logistic regression on pre-extracted OpenSMILE acoustic features, and the relevance model is based on MatchRNN, as we used for our text encoder.
\end{itemize}

\begin{figure}[!t]
\centering
\includegraphics[width=14cm]{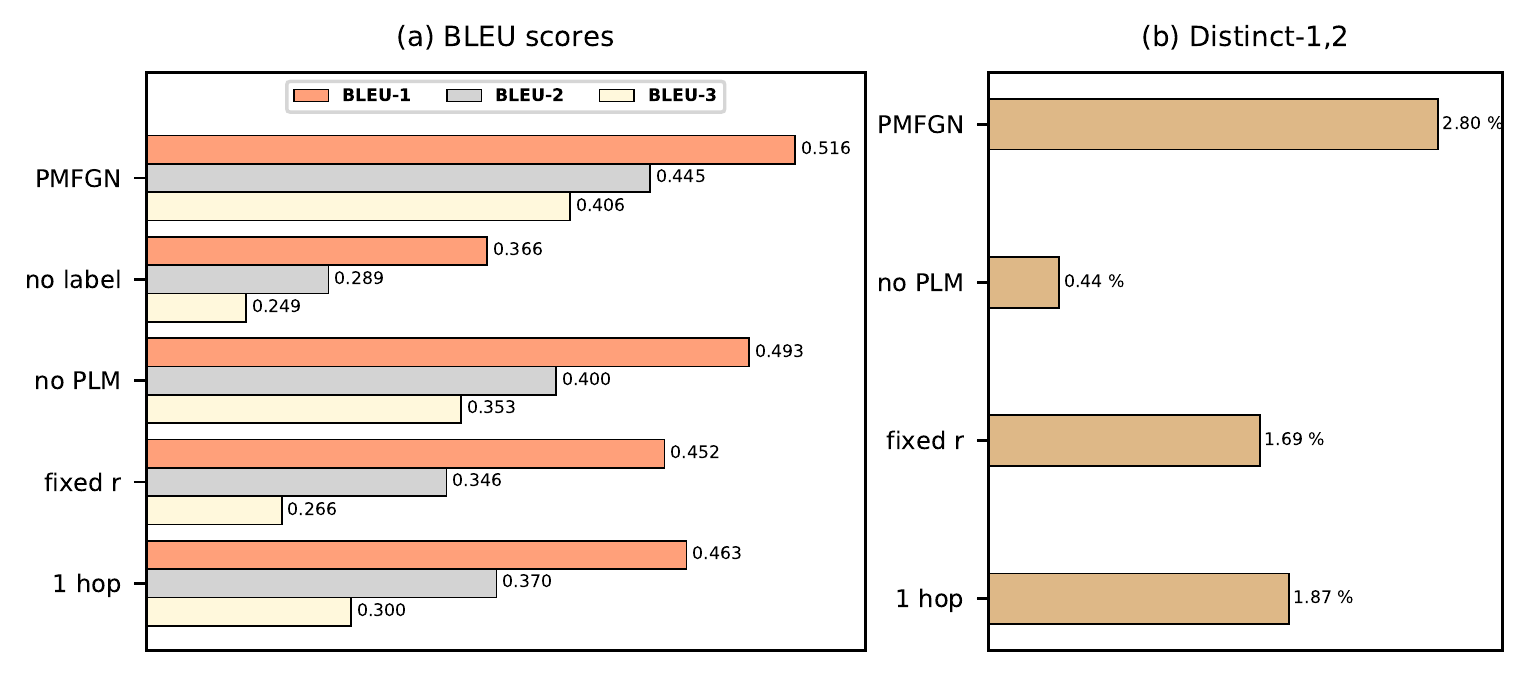}
\caption{The results of ablation study. PMFGN represents the entire model; ``no label'' stands for the model without label information for multi-task learning; ``no PLM'' is the model without the personalized language model; ``fixed r'' stands for that r is set to a fixed value of 0.7; ``1 hop'' represents that 1 hop structured self-attention is conducted in the modality gate.}
\label{fig:ablation}
\end{figure}

\subsection{Experimental Results}
We first evaluate the performance of our model and the baselines on the test data according to several automatic metrics: perplexity, BLEU, ROUGE, and Distinct-N. Perplexity measures how well a probabilistic generative model predicts the ground truth. The lower, the better. BLEU and ROUGE measure the text similarity between the generated feedback and reference feedback by comparing the number of overlapping text units. Besides the ground truth, we manually select 1-4 pieces of semantically identical feedback written by the same teacher as the references. Published evaluation codes are used for these two metrics\footnote{https://github.com/yunjey/show-attend-and-tell/blob/master/core/bleu.py}. Distinct-N measures the diversity of the generated texts by computing the number of distinct N-grams of the texts \cite{DBLP:conf/naacl/LiGBGD16}. A higher Distinct-N score indicates a more diverse text.

Experimental results are shown in Table \ref{tab:expresults}. From the table, we make the following observations: First, Multimodal Attention achieves higher BLEU and ROUGE scores than the first four methods. It demonstrates that the information from all the three modalities is useful for generating the feedback, and enables the model to make a more accurate evaluation, which is reflected by producing feedback more similar to the references. Second, all the baselines get relatively low Distinct scores, which means that without a personalized mechanism, these methods can only generate universal and repetitive feedback for all the teachers. Except for GRU-LM, which gets a higher Distinct score thanks to randomly sampling the first token. However, the generated feedback is indeed random too. Third, the traditional pipeline method Repository cannot obtain the desired performance, since the limited repositories are not applicable for all the cases and all the teachers. Finally, our model achieves the best results on all the evaluation metrics, which demonstrates that our model effectively takes advantage of the information of each modality to produce more proper feedback. Besides, through the personalized bias mechanism, our model generates feedback that is more diverse and in line with the style of teachers.

To further demonstrate the effectiveness of the proposed framework, we present a case study in Appendix \ref{appendix}. We provide three examples of oral presentation assignments along with the corresponding produced feedback. A discussion on the examples is also included.

\subsection{Ablation Study}
In this subsection, we seek to figure out how each component contributes to the final performance of the proposed model. In Figure \ref{fig:ablation}, we compare the performance of the entire proposed model and variants of the models with some components eliminated.

We make the following observations. First, if we don't use the annotated modality labels to train the model to predict the targeted modalities but randomly select modalities (i.e., ``no label'' in the figure), the performance on BLEU scores degrades dramatically, which demonstrates that choosing the right modalities is crucial for generating the correct feedback. Second, if we remove the personalized language model from the framework (i.e., ``no PLM" in the figure), the BLEU scores also decline and the Distinct score changes to $1/6$ of the original, which shows that the personalized bias mechanism helps a lot for generating personalized and more diverse feedback. Note that for the model without personalized language model, the BLEU scores are still higher than the scores of the baseline Multimodal Attention, which means our model is more effective to integrate multimodal information for feedback generation than directly employing a hierarchical attention mechanism for multimodal information fusion. Third, if the weight $r$ is fixed to be a constant instead of being determined based on the hidden state of the general language model at each step (i.e., ``fixed r" in the figure), the performances become even poorer than no PLM. It is because we need to let the model decide whether a ``general'' word or a ``personalized'' word should be generated based on the context. Otherwise, with a fixed weight, the model has constant biases on the two sides at each step, which sometimes leads to a conflict between the two language models in deciding which words to choose, so that ungrammatical sentences can be generated. Fourth, in the complete model, we adopt $K$-hop structured self-attention to extract the information of various aspects from a modality. If we replace multiple hops with 1 hop (i.e., ``1 hop" in the figure), the performance is also poorer than the original one, which verifies the benefits of this design.
\section{Related Work}
\label{sec:rela}

To the best of our knowledge, there is no previous work on the problem of automatic feedback generation for oral presentation assignments. However, there exist related works about text generation based on unimodal or multimodal information. The works \cite{DBLP:journals/corr/MaoXYWY14a,DBLP:conf/cvpr/VinyalsTBE15,DBLP:conf/icml/XuBKCCSZB15} investigate the problem of image captioning, which takes an image as input to generate a textual description. Video description is studied in the works \cite{venugopalan2014translating,rohrbach2015long} and \cite{DBLP:conf/iccv/HoriHLZHHMS17}. The first two take only dynamic images as input while the last one integrates the images and audio via a hierarchical attention mechanism to generate the description.

The works more similar to our work in the application are those which generate reviews for products on online shopping platforms. In the work \cite{DBLP:conf/eacl/ZhouLWDHX17}, a seq2seq model is adopted to encode attributes (user, product, rating) into vectors and then generate the review with an RNN decoder. An attention layer is added between encoder and decoder to learn the alignments between attributes and generated words. The work \cite{ni2018personalized} generates reviews for a given user, item, several related phrases such as product titles, and aspect-aware knowledge. In the work \cite{sun2019image}, the authors use an image of a product and a predetermined rating as evidence, to generate a review. For a given pair of user and item, the work \cite{DBLP:conf/www/TruongL19} first predicts a rating that the user would give to the item, and then integrate the multimodal information of the user, the item, the predicted rating, and an image to generate a review. In Net2Text \cite{xu2019net2text}, the authors first construct a graph with users and items as nodes and then predict a review of a user towards an item in a language model conditioned on learned node embeddings.

Personalized text generation is also a hot problem in many NLP tasks, especially dialogue generation. The most common solution is to introduce persona embeddings to model the personality of speakers \cite{DBLP:conf/acl/LiGBSGD16}. Besides, the work \cite{DBLP:conf/ijcnlp/LuanBDGG17} proposes to build personalized dialogue models via multi-task learning. The authors train a seq2seq model on common conversation data and an autoencoder on personal non-conversation data with the parameters in their decoders sharing. In addition, domain adaptation \cite{DBLP:journals/nn/YangTQZCZ18,DBLP:journals/www/ZhangZWZL19} and transfer learning \cite{DBLP:conf/aaai/MoZLLY18} methods are also proposed for personalized dialogue generation.
\section{Conclusion}
\label{sec:con}
In this paper, we study the problem of automated multimodal feedback generation for oral presentation assignments in K-12 education. As a pioneering work for this task, we propose a novel PMFGN that learns to produce personalized feedback for oral presentations in an end-to-end manner. Equipped with a modality gate mechanism and a personalized bias mechanism, the proposed framework encodes and fuses multimodal information in an effective way and achieves personalized feedback generation. The performance of the proposed model is demonstrated by experiments conducted on a real-world K-12 education dataset according to various evaluation metrics.

This work focuses on evaluating an oral presentation assignment based on its intrinsic quality (clarity, fluency, relevance, etc.), but doesn't consider judging whether an answer is right or wrong since the automated correction of assignments is another problem under study. In the future, we plan to integrate this aspect into feedback generation.

\section*{Acknowledgements}

Haochen Liu and Jiliang Tang are supported by the National Science Foundation of the United States under CNS1815636, IIS1928278, IIS1714741, IIS1845081, IIS1907704, and IIS1955285. Zitao Liu is supported by the Beijing Nova Program (Z201100006820068) from Beijing Municipal Science \& Technology Commission.

\bibliographystyle{coling}
\bibliography{main}

\begin{thebibliography}{}

\bibitem[\protect\citename{Baltrusaitis \bgroup et al.\egroup
  }2019]{DBLP:journals/pami/BaltrusaitisAM19}
Tadas Baltrusaitis, Chaitanya Ahuja, and Louis{-}Philippe Morency.
\newblock 2019.
\newblock Multimodal machine learning: {A} survey and taxonomy.
\newblock {\em {IEEE} Trans. Pattern Anal. Mach. Intell.}, 41(2):423--443.

\bibitem[\protect\citename{Caruana}1993]{DBLP:conf/icml/Caruana93}
Rich Caruana.
\newblock 1993.
\newblock Multitask learning: {A} knowledge-based source of inductive bias.
\newblock In {\em Machine Learning, Proceedings of the Tenth International
  Conference, University of Massachusetts, Amherst, MA, USA, June 27-29, 1993},
  pages 41--48.

\bibitem[\protect\citename{Hori \bgroup et al.\egroup
  }2017]{DBLP:conf/iccv/HoriHLZHHMS17}
Chiori Hori, Takaaki Hori, Teng{-}Yok Lee, Ziming Zhang, Bret Harsham, John~R.
  Hershey, Tim~K. Marks, and Kazuhiko Sumi.
\newblock 2017.
\newblock Attention-based multimodal fusion for video description.
\newblock In {\em {IEEE} International Conference on Computer Vision, {ICCV}
  2017, Venice, Italy, October 22-29, 2017}, pages 4203--4212.

\bibitem[\protect\citename{Kingma and Ba}2014]{kingma2014adam}
Diederik~P Kingma and Jimmy Ba.
\newblock 2014.
\newblock Adam: A method for stochastic optimization.
\newblock {\em arXiv preprint arXiv:1412.6980}.

\bibitem[\protect\citename{Kiros \bgroup et al.\egroup
  }2014]{DBLP:journals/corr/KirosSZ14}
Ryan Kiros, Ruslan Salakhutdinov, and Richard~S. Zemel.
\newblock 2014.
\newblock Unifying visual-semantic embeddings with multimodal neural language
  models.
\newblock {\em CoRR}, abs/1411.2539.

\bibitem[\protect\citename{Li \bgroup et al.\egroup
  }2016a]{DBLP:conf/naacl/LiGBGD16}
Jiwei Li, Michel Galley, Chris Brockett, Jianfeng Gao, and Bill Dolan.
\newblock 2016a.
\newblock A diversity-promoting objective function for neural conversation
  models.
\newblock In {\em {NAACL} {HLT} 2016, The 2016 Conference of the North American
  Chapter of the Association for Computational Linguistics: Human Language
  Technologies, San Diego California, USA, June 12-17, 2016}, pages 110--119.

\bibitem[\protect\citename{Li \bgroup et al.\egroup
  }2016b]{DBLP:conf/acl/LiGBSGD16}
Jiwei Li, Michel Galley, Chris Brockett, Georgios~P. Spithourakis, Jianfeng
  Gao, and William~B. Dolan.
\newblock 2016b.
\newblock A persona-based neural conversation model.
\newblock In {\em Proceedings of the 54th Annual Meeting of the Association for
  Computational Linguistics, {ACL} 2016, August 7-12, 2016, Berlin, Germany,
  Volume 1: Long Papers}.

\bibitem[\protect\citename{Lin \bgroup et al.\egroup
  }2017]{DBLP:conf/iclr/LinFSYXZB17}
Zhouhan Lin, Minwei Feng, C{\'{\i}}cero~Nogueira dos Santos, Mo~Yu, Bing Xiang,
  Bowen Zhou, and Yoshua Bengio.
\newblock 2017.
\newblock A structured self-attentive sentence embedding.
\newblock In {\em 5th International Conference on Learning Representations,
  {ICLR} 2017, Toulon, France, April 24-26, 2017, Conference Track
  Proceedings}.

\bibitem[\protect\citename{Liu \bgroup et al.\egroup }2019]{liu2019automatic}
Tiaoqiao Liu, Wenbiao Ding, Zhiwei Wang, Jiliang Tang, Gale~Yan Huang, and
  Zitao Liu.
\newblock 2019.
\newblock Automatic short answer grading via multiway attention networks.
\newblock In {\em International Conference on Artificial Intelligence in
  Education}, pages 169--173. Springer.

\bibitem[\protect\citename{Liu \bgroup et al.\egroup }2020a]{liu2020dolphin}
Zitao Liu, Guowei Xu, Tianqiao Liu, Weiping Fu, Yubi Qi, Wenbiao Ding, Yujia
  Song, Chaoyou Guo, Cong Kong, Songfan Yang, et~al.
\newblock 2020a.
\newblock Dolphin: A spoken language proficiency assessment system for
  elementary education.
\newblock In {\em Proceedings of The Web Conference 2020}, pages 2641--2647.

\bibitem[\protect\citename{Liu \bgroup et al.\egroup }2020b]{liu2020recent}
Zitao Liu, Songfan Yang, Jiliang Tang, Neil Heffernan, and Rose Luckin.
\newblock 2020b.
\newblock Recent advances in multimodal educational data mining in k-12
  education.
\newblock In {\em Proceedings of the 26th ACM SIGKDD International Conference
  on Knowledge Discovery \& Data Mining}, pages 3549--3550.

\bibitem[\protect\citename{Luan \bgroup et al.\egroup
  }2017]{DBLP:conf/ijcnlp/LuanBDGG17}
Yi~Luan, Chris Brockett, Bill Dolan, Jianfeng Gao, and Michel Galley.
\newblock 2017.
\newblock Multi-task learning for speaker-role adaptation in neural
  conversation models.
\newblock {\em arXiv preprint arXiv:1710.07388}.

\bibitem[\protect\citename{Mao \bgroup et al.\egroup
  }2015]{DBLP:journals/corr/MaoXYWY14a}
Junhua Mao, Wei Xu, Yi~Yang, Jiang Wang, and Alan~L. Yuille.
\newblock 2015.
\newblock Deep captioning with multimodal recurrent neural networks (m-rnn).
\newblock In {\em 3rd International Conference on Learning Representations,
  {ICLR} 2015, San Diego, CA, USA, May 7-9, 2015, Conference Track
  Proceedings}.

\bibitem[\protect\citename{Mo \bgroup et al.\egroup
  }2018]{DBLP:conf/aaai/MoZLLY18}
Kaixiang Mo, Yu~Zhang, Shuangyin Li, Jiajun Li, and Qiang Yang.
\newblock 2018.
\newblock Personalizing a dialogue system with transfer reinforcement learning.
\newblock In {\em Proceedings of the Thirty-Second {AAAI} Conference on
  Artificial Intelligence, (AAAI-18), the 30th innovative Applications of
  Artificial Intelligence (IAAI-18), and the 8th {AAAI} Symposium on
  Educational Advances in Artificial Intelligence (EAAI-18), New Orleans,
  Louisiana, USA, February 2-7, 2018}, pages 5317--5324.

\bibitem[\protect\citename{Ni and McAuley}2018]{ni2018personalized}
Jianmo Ni and Julian McAuley.
\newblock 2018.
\newblock Personalized review generation by expanding phrases and attending on
  aspect-aware representations.
\newblock In {\em Proceedings of the 56th Annual Meeting of the Association for
  Computational Linguistics (Volume 2: Short Papers)}, pages 706--711.

\bibitem[\protect\citename{Rohrbach \bgroup et al.\egroup
  }2015]{rohrbach2015long}
Anna Rohrbach, Marcus Rohrbach, and Bernt Schiele.
\newblock 2015.
\newblock The long-short story of movie description.
\newblock In {\em German conference on pattern recognition}, pages 209--221.
  Springer.

\bibitem[\protect\citename{Sun \bgroup et al.\egroup }2019]{sun2019image}
Xuehui Sun, Zihan Zhou, and Yuda Fan.
\newblock 2019.
\newblock Image based review text generation with emotional guidance.
\newblock {\em arXiv preprint arXiv:1901.04140}.

\bibitem[\protect\citename{Szegedy \bgroup et al.\egroup
  }2015]{DBLP:conf/cvpr/SzegedyLJSRAEVR15}
Christian Szegedy, Wei Liu, Yangqing Jia, Pierre Sermanet, Scott~E. Reed,
  Dragomir Anguelov, Dumitru Erhan, Vincent Vanhoucke, and Andrew Rabinovich.
\newblock 2015.
\newblock Going deeper with convolutions.
\newblock In {\em {IEEE} Conference on Computer Vision and Pattern Recognition,
  {CVPR} 2015, Boston, MA, USA, June 7-12, 2015}, pages 1--9.

\bibitem[\protect\citename{Truong and Lauw}2019]{DBLP:conf/www/TruongL19}
Quoc{-}Tuan Truong and Hady~W. Lauw.
\newblock 2019.
\newblock Multimodal review generation for recommender systems.
\newblock In {\em The World Wide Web Conference, {WWW} 2019, San Francisco, CA,
  USA, May 13-17, 2019}, pages 1864--1874.

\bibitem[\protect\citename{Venugopalan \bgroup et al.\egroup
  }2014]{venugopalan2014translating}
Subhashini Venugopalan, Huijuan Xu, Jeff Donahue, Marcus Rohrbach, Raymond
  Mooney, and Kate Saenko.
\newblock 2014.
\newblock Translating videos to natural language using deep recurrent neural
  networks.
\newblock {\em arXiv preprint arXiv:1412.4729}.

\bibitem[\protect\citename{Vinyals \bgroup et al.\egroup
  }2015]{DBLP:conf/cvpr/VinyalsTBE15}
Oriol Vinyals, Alexander Toshev, Samy Bengio, and Dumitru Erhan.
\newblock 2015.
\newblock Show and tell: {A} neural image caption generator.
\newblock In {\em {IEEE} Conference on Computer Vision and Pattern Recognition,
  {CVPR} 2015, Boston, MA, USA, June 7-12, 2015}, pages 3156--3164.

\bibitem[\protect\citename{Wang and Jiang}2016]{DBLP:conf/naacl/WangJ16}
Shuohang Wang and Jing Jiang.
\newblock 2016.
\newblock Learning natural language inference with {LSTM}.
\newblock In {\em {NAACL} {HLT} 2016, The 2016 Conference of the North American
  Chapter of the Association for Computational Linguistics: Human Language
  Technologies, San Diego California, USA, June 12-17, 2016}, pages 1442--1451.

\bibitem[\protect\citename{Xu \bgroup et al.\egroup
  }2015]{DBLP:conf/icml/XuBKCCSZB15}
Kelvin Xu, Jimmy Ba, Ryan Kiros, Kyunghyun Cho, Aaron~C. Courville, Ruslan
  Salakhutdinov, Richard~S. Zemel, and Yoshua Bengio.
\newblock 2015.
\newblock Show, attend and tell: Neural image caption generation with visual
  attention.
\newblock In {\em Proceedings of the 32nd International Conference on Machine
  Learning, {ICML} 2015, Lille, France, 6-11 July 2015}, pages 2048--2057.

\bibitem[\protect\citename{Xu \bgroup et al.\egroup }2019]{xu2019net2text}
Shaofeng Xu, Yun Xiong, Xiangnan Kong, and Yangyong Zhu.
\newblock 2019.
\newblock Net2text: An edge labelling language model for personalized review
  generation.
\newblock In {\em International Conference on Database Systems for Advanced
  Applications}, pages 484--500. Springer.

\bibitem[\protect\citename{Yang \bgroup et al.\egroup
  }2018]{DBLP:journals/nn/YangTQZCZ18}
Min Yang, Wenting Tu, Qiang Qu, Zhou Zhao, Xiaojun Chen, and Jia Zhu.
\newblock 2018.
\newblock Personalized response generation by dual-learning based domain
  adaptation.
\newblock {\em Neural Networks}, 103:72--82.

\bibitem[\protect\citename{Zhang \bgroup et al.\egroup
  }2019]{DBLP:journals/www/ZhangZWZL19}
Weinan Zhang, Qingfu Zhu, Yifa Wang, Yanyan Zhao, and Ting Liu.
\newblock 2019.
\newblock Neural personalized response generation as domain adaptation.
\newblock {\em World Wide Web}, 22(4):1427--1446.

\bibitem[\protect\citename{Zhao and
  Esk{\'{e}}nazi}2016]{DBLP:journals/corr/ZhaoE16}
Tiancheng Zhao and Maxine Esk{\'{e}}nazi.
\newblock 2016.
\newblock Towards end-to-end learning for dialog state tracking and management
  using deep reinforcement learning.
\newblock {\em CoRR}, abs/1606.02560.

\bibitem[\protect\citename{Zhou \bgroup et al.\egroup
  }2017]{DBLP:conf/eacl/ZhouLWDHX17}
Ming Zhou, Mirella Lapata, Furu Wei, Li~Dong, Shaohan Huang, and Ke~Xu.
\newblock 2017.
\newblock Learning to generate product reviews from attributes.
\newblock In {\em Proceedings of the 15th Conference of the European Chapter of
  the Association for Computational Linguistics, {EACL} 2017, Valencia, Spain,
  April 3-7, 2017, Volume 1: Long Papers}, pages 623--632.

\end{thebibliography}

\appendix
\section{Appendix A. Case Study}
\label{appendix}


In the appendix, we show three oral presentation assignment examples and their corresponding submissions and generated feedback to understand the superiority of our proposed method. Note that the original examples are in Chinese and we translate them into English. Following the examples, we provide a discussion on them in Section \ref{sec:discuss}.

\begin{CJK*}{UTF8}{gbsn}
\subsection{Example A}

\noindent\textbf{Question}: 根据今天学到的知识，我们知道汉字图案“田”是不能一笔画出来的．那么说说可以进行怎样的修改，变成能一笔画出的图案？（要求：不得减少或增加交点）


\noindent(\emph{Translation: } According to the knowledge we have learned today, we know that the Chinese character ``田'' can not be drawn in one stroke.
So what modifications can be made to turn it into a pattern that can be drawn in one stroke? (Requirement: You cannot decrease or increase the number of intersection points)

\noindent\textbf{Submitted video (screenshot)}:

\begin{figure}[H]
\centering
\includegraphics[width=8cm]{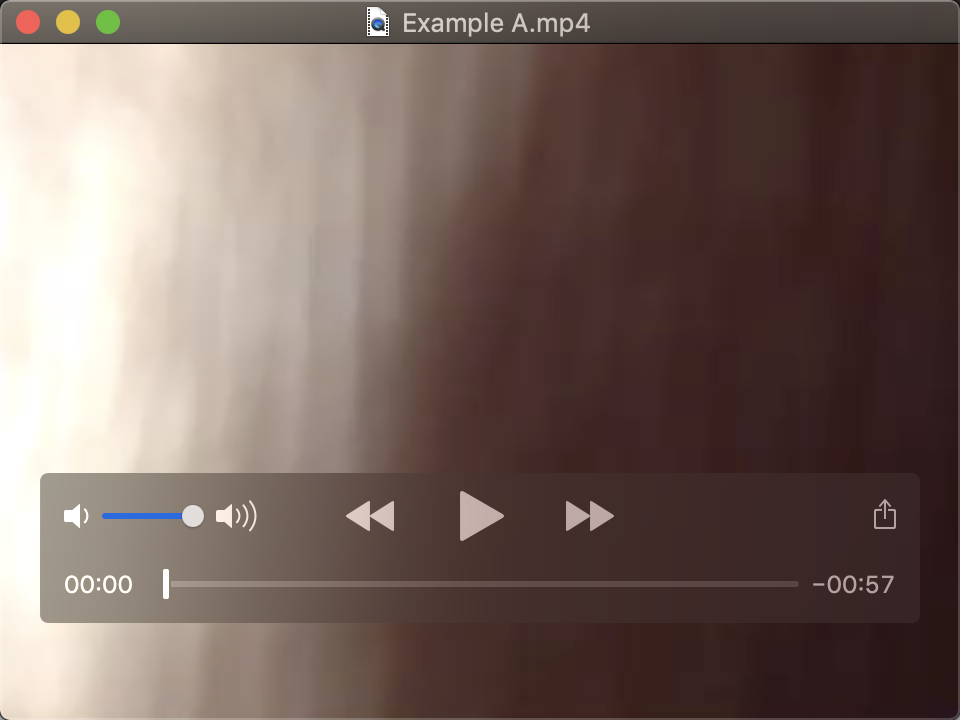}
\end{figure}

\noindent The feedback written by a teacher, produced by baseline methods and our proposed model is listed as follows.

\noindent\textbf{Teacher \#97: 镜头没有对好哦}

\noindent(\emph{Translation:}  \textbf{The camera is not aimed properly})

\noindent\textbf{GRM-LM}: 黑屏？

\noindent(\emph{Translation: }  Black screen?)

\noindent\textbf{Show-Attend-and-Tell}: 非常棒！

\noindent(\emph{Translation: }  Very good!)

\noindent\textbf{Attribute2Seq-Audio}: 非常棒

\noindent(\emph{Translation: }  Very good)

\noindent\textbf{Attribute2Seq-Text}: 非常棒，继续加油

\noindent(\emph{Translation: }  Very good, keep going)

\noindent\textbf{Multimodal Attention}: 非常棒

\noindent(\emph{Translation: }  Very good)

\noindent\textbf{Repository}: 准备得很用心，给你点赞！表达很流畅，表现力也很强。下次也要继续保持认真的态度！

\noindent(\emph{Translation: }  It's prepared carefully. Give you a like! It's very fluent and expressive. We should continue to be serious the next time!)

\noindent\textbf{PMFGN(with the personality of the teacher \#97): 说的很对！但是老师看不见你哟}

\noindent(\emph{Translation: }  \textbf{What you said is correct! But I can not see you})

\subsection{Example B}

\noindent\textbf{Question}: 请你讲一讲，用什么方法才能既准确又快速地计算下列算式的末三位数呢？


\noindent(\emph{Translation:} Could you tell me how to calculate the last three digits of the following formula accurately and quickly?)

\begin{equation}
5+55+555+ \cdots +\underbrace{55 \cdots 5}_{20\ times}
\end{equation}

\noindent\textbf{Submitted video (screenshot)}:

\begin{figure}[H]
\centering
\includegraphics[width=8cm]{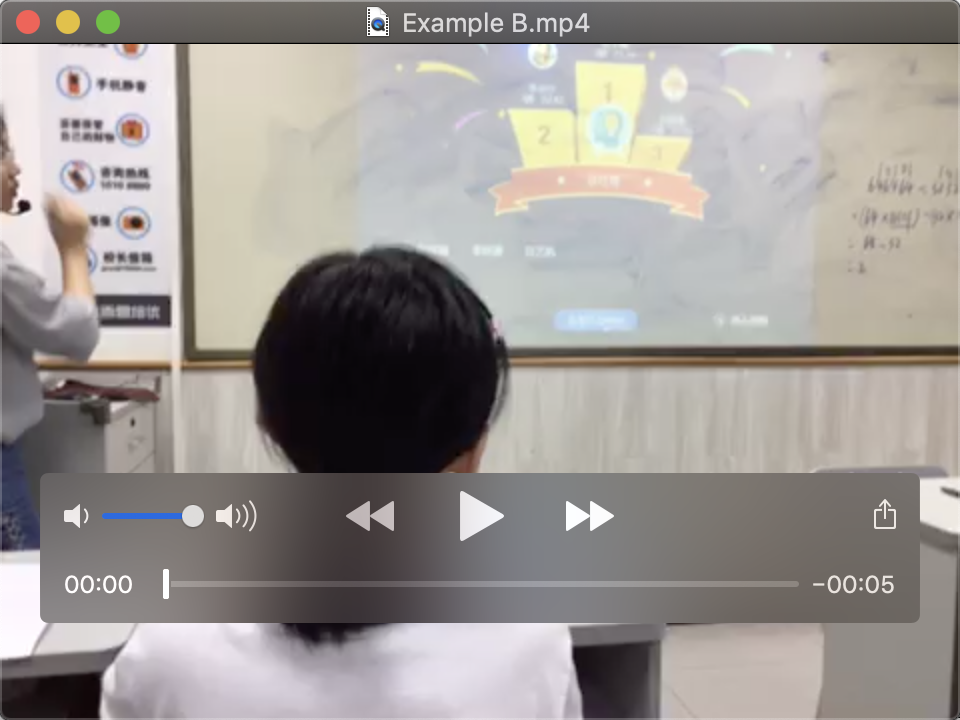}
\end{figure}

\noindent The feedback written by a teacher, produced by baseline methods and our proposed model is listed as follows.

\noindent\textbf{Teacher \#107: 是不是录错了呀！}

\noindent(\emph{Translation: }  \textbf{Did you record a wrong video!})

\noindent\textbf{GRM-LM}: 不错，继续加油

\noindent(\emph{Translation: }  Good, keep going)

\noindent\textbf{Show-Attend-and-Tell}: 非常棒！

\noindent(\emph{Translation: }  Very good!)

\noindent\textbf{Attribute2Seq-Audio}: 录制的和这讲内容无关哦

\noindent(\emph{Translation: }  What you record has nothing to do with the content of this section)

\noindent\textbf{Attribute2Seq-Text}: 视频没有声音呢？

\noindent(\emph{Translation: }  There is no sound in the video?)

\noindent\textbf{Multimodal Attention}: 看不到

\noindent(\emph{Translation: }  I can’t see)

\noindent\textbf{Repository}: 录制的和这讲内容无关诶

\noindent(\emph{Translation: }  What you record has nothing to do with the content of this section)

\noindent\textbf{PMFGN(with the personality of the teacher \#107): 是不是录错了呀宝贝}

\noindent(\emph{Translation: }  \textbf{Did you record a wrong video, baby})

\subsection{Example C}

\noindent\textbf{Question}: 迈克和南希一共有80块糖果，迈克给南希2块糖果之后，两人糖果数一样多。你知道他们原来各有多少块糖吗?

\noindent(\emph{Translation:} Mike and Nancy had 80 pieces of candy. After Mike gave Nancy 2 pieces of candy, they have the same number of candies. Do you know how many pieces of candy they had respectively at the beginning?)

\noindent\textbf{Submitted video (screenshot)}:

\begin{figure}[H]
\centering
\includegraphics[width=8cm]{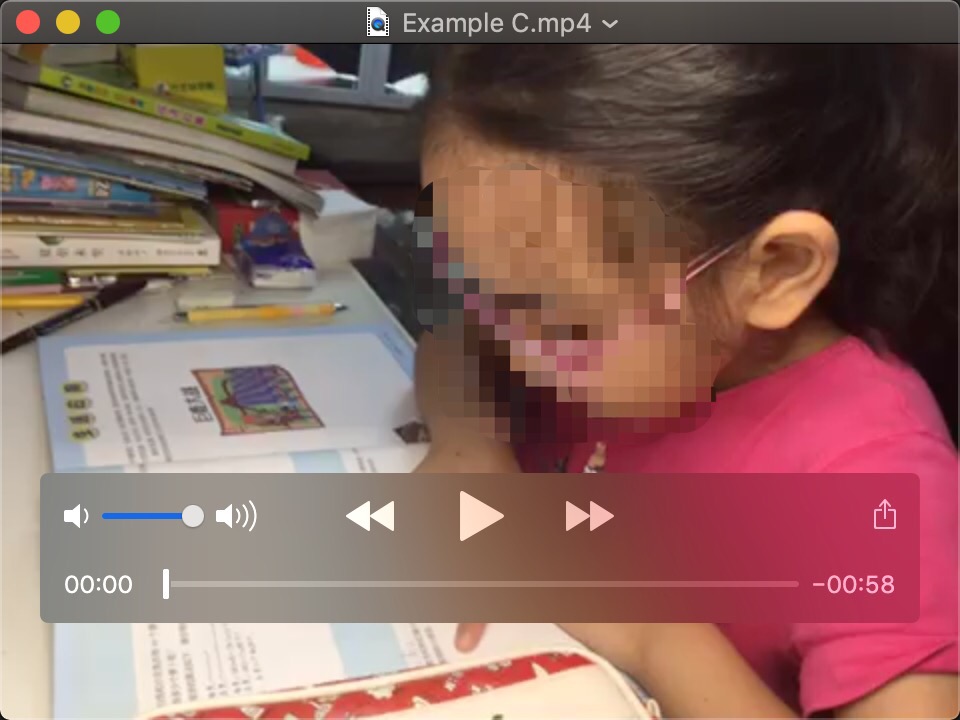}
\end{figure}

\noindent The feedback written by a teacher, produced by baseline methods and our proposed model is listed as follows.

\noindent\textbf{Teacher \#64: 思路清楚，表达流利！赞}

\noindent(\emph{Translation: } \textbf{Your mind is clear, and you express it fluently! Awesome})

\noindent\textbf{GRM-LM}: 黑屏？

\noindent(\emph{Translation: } Black screen?)

\noindent\textbf{Show-Attend-and-Tell}: 非常棒！

\noindent(\emph{Translation: } Very good!)

\noindent\textbf{Attribute2Seq-Audio}: 非常棒

\noindent(\emph{Translation: } Very good)

\noindent\textbf{Attribute2Seq-Text}: 非常棒！继续加油！

\noindent(\emph{Translation: } Very good! Please keep going!)

\noindent\textbf{Multimodal Attention}: 非常好！

\noindent(\emph{Translation: } Very good!)

\noindent\textbf{Repository}: 完成得不错噢，准备的比较认真。如果再流利一点就更棒啦。期待你下次的表现，加油！

\noindent(\emph{Translation: } Well done. Well prepared. It would be better if you were a little more fluent. Looking forward to your performance in the next time, come on!)

\noindent\textbf{PMFGN(with the personality of the teacher \#64): 非常棒，思路清晰，表达流利！完美！}

\noindent(\emph{Translation: } \textbf{Very good, your mind is clear, and you express it fluently! Perfect!})

\end{CJK*}

\subsection{Discussion}
\label{sec:discuss}
Example A shows an oral presentation submission with a video of 57 seconds long. The sound of the video is very clear and the answer is correct, but the image is very blurred. The video is not shot aimed at the student as required. Except for GRM-LM, all the baselines fail to recognize the defect on the image, and provide completely positive feedback. Without the image information as input, GRM-LM generates a piece of feedback ``Black screen?'' randomly. Our method produces a piece of perfect feedback that admires the student's answer and also points out the issues on the image. For example B, a wrong video with a length of only 5 seconds is submitted. Our model not only accurately points out this error, but also produces a piece of feedback which is very similar to the feedback given by the teacher. This means that our model successfully imitates the teacher's style and generates personalized feedback. Example C shows an eligible oral presentation submission. Although most of the baselines produce general positive feedback, only our method generates a piece of specific feedback which gives a positive comment on the fluency of the voice, and at the same time, it is consistent with the teacher's feedback.

\end{document}